\documentclass[journal]{IEEEtran}
\usepackage{epsf}
\usepackage{amsmath,amssymb}
\usepackage{graphicx}
\usepackage{color}
\usepackage{cite}
\usepackage{multirow,tabularx}
\usepackage{ifthen}
\usepackage{times}
\usepackage{stackrel}
\usepackage[hidelinks]{hyperref}
\usepackage{enumerate}
\usepackage{mathtools}

\usepackage{algorithm}
\usepackage{algorithmic}

\graphicspath{{./plots/}}

\usepackage[caption=false]{subfig}
\usepackage{tabularx}
\usepackage{ragged2e}

\IEEEoverridecommandlockouts

\begin{document}
	
	
\title{A UAV-Based Multispectral and RGB Dataset for Multi-Stage Paddy Crop Monitoring in Indian Agricultural Fields }

\author{ Adari Rama Sukanya, Puvvula Roopesh Naga Sri Sai, Bodduru
	Neshika, \\
	 Rimalapudi Sarvendranath, {\it Senior Member, IEEE},  
\thanks{
	Emails:adarisukanya2003@gmail.com, puvvularoopesh2004@gmail.com, moseswesly06@gmail.com, sarvendranath@gmail.com.    
	This dataset was jointly collected by IIT Tirupati and DroneHub Technologies Pvt. Ltd.  with funding support from IIT Tirupati Navavishkar I-Hub Foundation (IITTNiF) under the project grant IITTNiF/TPD/2024-25/P18.}

}
\maketitle
		
\begin{abstract}

We present a large-scale unmanned aerial vehicle (UAV)-based RGB and multispectral image dataset collected over paddy fields in the Vijayawada region, Andhra Pradesh, India, covering nursery to harvesting stages. We used a 20-megapixel RGB camera and a 5-megapixel four-band multispectral camera capturing red, green, red-edge, and near-infrared bands. Standardised operating procedure (SOP) and checklists were developed to ensure repeatable data acquisition. 
Our dataset comprises of 42,430 raw images (415 GB)  captured over 5 acres with 1 cm/pixel ground sampling distance (GSD) with associated metadata such as GPS coordinates, flight altitude, and environmental conditions. Captured images were validated using Pix4D Fields to generate orthomosaic maps and vegetation index maps, such as normalised difference vegetation index (NDVI) and normalised difference red-edge (NDRE) index. Our dataset is one of the few datasets that provide high-resolution images with rich metadata that cover all growth stages of Indian paddy crops. 
The dataset is available on IEEE DataPort with DOI,  .
It can support studies on targeted spraying, disease analysis, and yield estimation.

\end{abstract}


\section{Introduction}	
\label{sec:intro}

In the recent years, unmanned aerial vehicles (UAV) based imaging is shown to be promising to perform systematic observation, measurement, and assessment of crop conditions and field parameters. 
They immensely support precision farming and helps in increasing the  yield~\cite{Rao_2018_PCS}. 
The development of compact and advanced multispectral cameras that capture  data at various discrete wavelengths across the electromagnetic spectrum, has significantly expanded the scope and reliability of aerial crop imaging \cite{Tsouros_2019_info}.
The images captured by the UAV, enable detailed assessment of crop health, growth patterns, field variability. 
Besides, they also help in the early detection of stress factors such as nutrient deficiencies, water scarcity,  pest infestations, as well as discriminating between healthy and unhealthy vegetation, and also bare soil~\cite{Hegarty_2020_Sensing, MDPI_2025_Drones,Bukowiecki_2021_Sensors}.
Vegetation indices, such as the normalized difference vegetation index (NDVI), which is calculated using the reflectance value of red and near-infrared (NIR) bands, are widely used to quantify vegetation vigor and overall canopy condition. 
The normalized difference red edge (NDRE), derived from the reflectance value of red-edge (RE) and NIR bands, is more sensitive to subtle chlorophyll variations and early-stage vegetation stress. 
Unlike satellite imagery, UAVs provide greater temporal flexibility, they can capture data based on need to match crop growth cycles and respond to sudden changes. 
This combination of precision, adaptability, and timely data acquisition makes UAV based crop health monitoring an effective and reliable solution for enhanced crop monitoring practices.
Furthermore, UAV imagery provides a higher ground sampling distance (GSD), which is defined as the distance between the centers of two adjacent pixels on the ground, representing the real area covered by each pixel in the image. 
UAVs can capture images with a GSD of the order of centimeters compared to the satellites images whose GSD is of the order of meters.


\subsection{RGB and Multispectral Imaging Datasets}

{\em UAV-Based Datasets:}
The existing UAV based datasets for agricultural applications are as follows:
In~\cite{6_Maulit_2023_Data}, a dataset of approximately 38,377 images was collected over a 10-day interval across 27 hectares in Eastern Kazakhstan for wheat, barley, and soybean crops, covering five key growth stages. 
In~\cite{7_Fonseka_2024_Brief}, a dataset of 416 images was collected over 0.06 hectares in Sri Lanka for paddy during vegetative, reproductive, and ripening stages. 
A total of 7,638 images of Theobroma cacao plantations was acquired in the Central-West region of Côte d'Ivoire, covering 30 hectares  during the growth stage~\cite{8_Sabine_2024_Brief}. 
A multispectral UAV dataset of spring-sown pea and chickpea with 275 images was collected in~\cite{Umani_2024_Pulse}. 
In~\cite{12_Muhindo_2025_Agrosystems}, multispectral images were collected over paddy fields in the Ruzizi Plain, Democratic Republic of the Congo, focusing on active tillering and panicle initiation stages.
United States Department of Agriculture - Agricultural Research Service (USDA-ARS)~\cite{13_Farag_2024_Phenome}, provides a paddy dataset of six spectral bands that cover emergence to boot stages across 224 and 152 plots in 2021 and 2022, respectively.
A paddy dataset that cn be used to assess the crop condition through vegetation indices and spectral reflectance of the region Jitra, Kedah, Malaysia is provided in~\cite{14_Mohidem_2022_IOP}.
For maize fields, a multispectral and multitemporal UAV dataset was collected in~\cite{WeedDataset_2025_arXiv}. It provides dense annotations for multiple weed species across growth stages and supports segmentation and weed-mapping tasks.
A paddy dataset for varieties INPARI-32, INPARI-33, and INPARI-43 were collected over a 90-meter-wide field in Indonesia at three growth stages after planting in~\cite{15_Wijayanto_2023_AgriEngineering}.
In \cite{10_Youness_2025_Brief, Ruwanpathirana_2024_Sugarcane_RGB, 11_Sayali_2025_Brief} collected only RGB UAV datasets for olive orchards, palm trees, and soybean crops, respectively.
 These datasets are summarized in Table~\ref{tab:crop_overview}.

\begin{table*}
	\centering
	
	\renewcommand{\arraystretch}{1.1}
	\setlength{\tabcolsep}{4pt}
	
	\caption{Comparison of Existing Datasets With Our Dataset}
	\label{tab:crop_overview}
	
	\begin{tabularx}{\textwidth}{
			>{\RaggedRight\arraybackslash}p{0.11\textwidth}
			>{\RaggedRight\arraybackslash}p{0.18\textwidth}
			>{\RaggedRight\arraybackslash}p{0.18\textwidth}
			>{\RaggedRight\arraybackslash}p{0.10\textwidth}
			>{\RaggedRight\arraybackslash}X}
		
		\hline
		Source & Location & Crop & Area (acres) & Stages of Crop \\
		\hline
		
		\cite{Bukowiecki_2021_Sensors} & Northern Germany &
		Winter wheat & 28.24 & — \\
		
		\cite{6_Maulit_2023_Data} & East Kazakhstan &
		Wheat, soybean, barley &
		66.72 &
		Emergence, tillering, heading, milky ripe and waxy ripe. \\
		
		\cite{7_Fonseka_2024_Brief} & Sri Lanka &
		BG300 (Paddy) &
		0.148 &
		Vegetative, reproductive and ripening. \\
		
		\cite{8_Sabine_2024_Brief} & Côte d'Ivoire &
		Cocoa trees &
		74.12 &
		Growing stage \\
		
		\cite{12_Muhindo_2025_Agrosystems} & Ruzizi Plain, DRC &
		Paddy &
		--- &
		Active tillering and panicle initiation. \\
		
		\cite{13_Farag_2024_Phenome} & US &
		Paddy &
		--- &
		Emergence to boot stage \\
		
		\cite{14_Mohidem_2022_IOP} & Jitra, Kedah, Malaysia &
		Paddy &
		--- &
		— \\
		
		\cite{15_Wijayanto_2023_AgriEngineering} & Indonesia &
		Paddy (INPARI-32, 33, 43) &
		90 m width &
		Vegetative and booting. \\
		
		\cite{17_Govi_2024_Sensing} & Italy &
		Vineyards &
		42.01 &
		— \\
		
		\cite{18_Richard_2017_Sensors} & Machakos, Kenya &
		Maize &
		167319 &
		Stem elongation and flowering. \\
		
		\cite{Sani_2024_SICKLE} & Tamil Nadu, India &
		Multi-crop & 898 &
		Sowing, transplanting, tillering, and harvesting. \\
		
		\cite{TiHAN_IITH_Paddy} & India &
		Paddy &
		--- &
		Weed, seedling, tillering, booting, flowering, and ripening. \\
		
		\cite{Sharma_2020_BasmatiPaddySeeds} & New Delhi India &
		Basmati paddy (seed-varieties) &
		--- &
		Seed-level variety imaging (no field growth stages). \\

		Our dataset & India &
		Paddy (MTU 13/18, 10/61) &
		5 &
		Nursery, vegetative, booting, flowering,\newline
		mature, and harvesting. \\
		
		\hline
	\end{tabularx}
\end{table*}

{\em Satellite Based Datasets:} Several satellite-based agricultural datasets have been reported in the literature. 
A Sentinel-2 based dataset for winter wheat monitoring was presented in~\cite{Bukowiecki_2021_Sensors}, covering 11.43 hectares in Northern Germany. 
In~\cite{17_Govi_2024_Sensing}, a dataset combining Sentinel-2 and UAV imagery was collected over 17 hectares of vineyard fields in Italy.
RapidEye satellite imagery was used in~\cite{18_Richard_2017_Sensors} to monitor maize cultivation across 677~km$^2$ in Machakos County, Kenya.
The EuroSAT dataset~\cite{EuroSAT_2018_Helber} consists of approximately 27,000 Sentinel-2 multispectral image patches spanning 13 spectral bands and covering 10 land-use and land-cover classes across Europe. 


Most existing publicly available crop-monitoring datasets remain limited in their ability to support comprehensive and reproducible analysis, particularly for paddy cultivation under Indian agro-climatic conditions. Specifically, these datasets often lack complete crop growth cycle coverage, paddy-specific standardized operating procedure documentation, access to raw RGB and multispectral imagery, and associated environmental metadata. Such limitations constrain longitudinal analysis, reproducibility, and region-specific modeling, thereby motivating the development of a comprehensive and well-documented UAV dataset focused on paddy crop monitoring.

\subsection{Our Contributions} 
\begin{itemize}
	\item We provide an SOP and flight checklists for the data acquisition methodology to ensure reproducibility	 and can be adapted to other UAV platforms.

	\item We present RGB and four-band (red, green, RE, NIR) multispectral image dataset of 5-acre paddy field for Indian agro-climatic conditions following the developed SOP and checklists. our dataset covers the complete growth cycle of the paddy crop and addresses the significant gap in the publicly available agricultural UAV datasets. 
	During capture, we ensured that 30–32 GNSS satellites were connected for reliable positional accuracy.
		
	\item 
	We provide 42,430 raw images (414~GB), covering all growth stages of the paddy crop. Our RGB images have a resolution of 20~-MP (5280 × 3956 pixels)  and multispectral images have a resolution of  5~-MP (2592 × 1944 pixels) with 1~cm/pixel GSD.

	\item Along with the dataset we also provide operational and environmental metadata, including captured location based on Global Navigation Satellite System (GNSS) coordinates, acquisition date and time, illumination conditions, solar elevation and azimuth angles, weather parameters such as temperature, humidity, and wind speed, camera intrinsic and extrinsic parameters, flight altitude, image overlap, GSD, and sensor calibration details.

	\item We validated the captured data through preprocessing steps such as  geometric alignment, band and radiometric calibrations. We also   georeferenced orthomosaic maps and vegetation indices (NDVI and NDRE).  
	
	\item  Our dataset enables a wide range of precision agriculture applications including targeted spraying, early pest and crop stress detection, disease analysis, crop classification, growth-stage monitoring, biomass and yield estimation, and temporal change analysis across growth stages.
	
	Our dataset is publicly released via IEEE DataPort with a persistent DOI,
	
		\end{itemize}

\subsection{Uniqueness of Our Dataset}
\begin{itemize}

\item 
Our dataset includes both RGB and multispectral images unlike datasets in~\cite{10_Youness_2025_Brief,Ruwanpathirana_2024_Sugarcane_RGB,11_Sayali_2025_Brief}, which provide only RGB images, and datasets in~\cite{6_Maulit_2023_Data,7_Fonseka_2024_Brief,Umani_2024_Pulse,13_Farag_2024_Phenome,15_Wijayanto_2023_AgriEngineering}, which provides only multispectral images. 
  	   	
\item 
Our dataset covers all six stages: nursery, vegetative, booting, flowering, mature, and harvest.
This is different from the other datasets that cover fewer stages as shown in the Table~\ref{tab:crop_overview}.

\item Our dataset offers higher spatial resolution, i.e.,  1~cm/pixel GSD,  compared to the existing datasets as shown in Table~\ref{tab:flight_comparison}, and allows accurate crop monitoring.
It also provides rich operational and environmental metadata. 

\item Our dataset provides SOP based paddy data for Indian agro-climatic conditions, addressing the gap of very limited datasets for India paddy crops.
Existing datasets for Indian crops either lack UAV images~\cite{Sani_2024_SICKLE} or multispectral data~\cite{Sharma_2020_BasmatiPaddySeeds}.
Our dataset provides high-resolution UAV based RGB and multispectral images with rich metadata across all six paddy growth stages, making it one of the unique datasets for Indian paddy crops.

     	

\end{itemize}

\subsection {Outline}

Section II describes the SOP and checklists developed for UAV-based RGB and multispectral data acquisition. Section III presents a detailed data description, including the acquisition site, data volume, nomenclature, and data organization. Section IV discusses the data quality and validation procedures using Pix4D software and highlights the potential use cases and applications of the proposed dataset. Section V concludes the paper.

\section{SOP and Checklists }

We developed SOP and checklists to ensure reprehensibility and can be adapted to other UAV platforms and sensing systems.
Using the developed SOP and checklists, we created a multispectral and RGB based UAV imaging dataset using a four 5 megapixel multispectral camera including Red, Green, RE, and NIR bands and an 20 megapixel RGB camera with image resolution of 2592 × 1944 pixels for multispectral images and 5280 × 3956 pixels for RGB images of different fields covering around 5 acres, where field 1 covers 3 acres and field 2 covers 2 acres. The dataset includes paddy crop varieties MTU 13/18 and MTU 10/61 cultivated during the Kharif Season and spans six different stages of the crop.

\begin{figure}
	\includegraphics[width=1\linewidth]{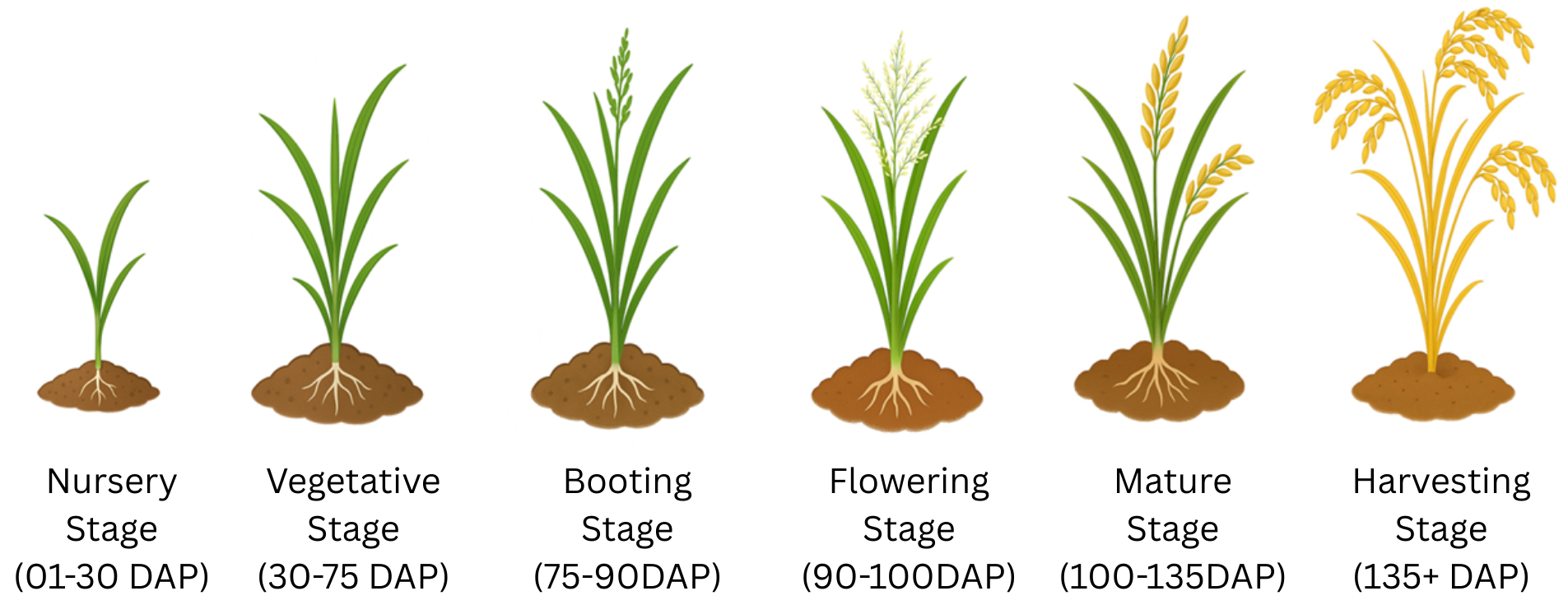}
	\caption{ Different stages of Paddy Crop}
	\label{fig:Stages}
\end{figure}

The nursery stage (1–30 days after plantation (DAP)) involves the planting of the seedlings with initial root and leaf growth. In the vegetative stage (30–75 DAP), rapid leaf development and canopy formation occur, and side shoots emerge, boosting the crop's yield potential. The booting stage (75–90 DAP) marks the formation of panicle inside the top leaf sheath. The flowering stage (90–100 DAP) is when panicles emerge and pollination takes place, which is crucial for grain development. In the mature stage (100–135 DAP), grain filling reaches completion and the plant begins to lose green color. Finally, in the harvest stage (post-135 DAP), the crop reaches full maturity and harvesting can begin. Figure~\ref{fig:Stages} describes different stages of the paddy crop.

A step-by-step SOP was developed to ensure consistent, accurate, and reliable  data collection. It outlines essential aspects such as airspace classification, climatic checks, mission planning, and optimal capture time in a day. This SOP also standardizes UAV configurations, flight checklists, and data management, providing a structured framework for reliable and high-quality outcomes. The overall checklist was separated into three parts: pre-flight, in-flight, and post-flight checklists. For ease of understanding we created data description, field weekly status reports and images. This formulated SOP is followed every time for the data capture  to generate an accurate dataset and avoid inconsistencies.
Our SOP has eight steps, which we explain in detail below.

\subsubsection{\text{Step 1: Check the Airspace Zones}}

Firstly, before every flight, check the official airspace map provided by your country’s aviation authority to identify the zone to which the area belongs and the corresponding flight regulations. This ensures operational safety and compliance with aviation regulations. For UAV operations, majorly classified into three zones: i) Red zone, which is no fly region, ii) Yellow zone, where controlled flights are allowed with prior permission, and iii) Green zone, where no permission is required to fly but within the permitted altitude limit.

\subsubsection{\text{Step 2: Check Weather Data}}

UAV data should be captured only under favorable weather conditions, avoiding rain, dense cloud cover that reduces image clarity, and intense sunlight that can cause overexposure. Wind speed should be below 5 m/s to ensure flight stability, and the UAV’s operational limits must be verified under the prevailing conditions. Global tools such as Windy can be used for wind forecasts, AccuWeather or Meteoblue for weather information, and SunCalc for assessing sun position and illumination angles.

\subsubsection{\text{Step 3:  Select Optimal Flight Timing}}

Plan UAV flights during balanced daylight, when the sun is at a moderate angle.  Always adjust flight timing based on local daylight patterns and seasonal variations to capture consistent, high-quality imagery. Avoid flying at midday, intense sunlight can cause glare, harsh shadows, and sensor overexposure. Choose times with stable temperatures and calm winds to prevent battery overheating and ensure flight stability.
\subsubsection{\text{Step 4: Perform Pre-Flight Checks}}
Before each UAV mission for data collection, pre-flight preparation is essential to ensure both flight safety and the quality of captured data. Below are the basic considerations,

\begin{itemize}
	\item \text{Check battery status:} Drone battery should be above 95\% with spares for the mission, and the controller battery should be above 40\% to ensure consistent communication between the drone and the controller.
	
	\item \text{Conduct physical inspection:} All components, including propellers, motors, and gimbal, must be inspected before the flight, and any damage should be fixed prior to the mission.
	
	\item \text{Ensure clean takeoff and landing area:} Use a flat, clear landing pad for takeoff and landing to avoid dust, dirt, or debris interfering with the sensors or propellers.
	
	\item \text{Use a high-speed SD card:} Insert a high-capacity, high-speed SD card with sufficient free space, as multispectral imaging generates large data volumes.
	
	\item \text{Calibrate with reflectance panels:} Place the panels under uniform lighting and capture reference images before and after the flight to ensure accurate radiometric calibration.
\end{itemize}
\subsubsection{\text{Step 5: Plan the Mission}}
Plan the mission carefully to ensure accurate data capture and safe UAV operation. Follow these key steps:
\begin{itemize}
	\item \text{Inspect field for obstacles:} Identify trees, power lines, and buildings, and plan the flight path to avoid them.
	
	\item \text{Set altitude and flight speed:} Adjust altitude and speed according to drone capability and mission requirements. Avoid flying too fast to prevent motion blur and too slow to maintain efficiency.
	
	\item \text{Adjust GSD:} Configure flight altitude to achieve the required image resolution. For crop stress detection, a general GSD of 1–3 ~cm per pixel is recommended.
	
	\item \text{Confirm stable GPS lock:} Ensure strong GPS connectivity before takeoff for accurate geotagging and navigation.
	
	\item \text{Set image overlap:} Use 80\% forward overlap and 70\% side overlap to provide redundancy for precise orthomosaic stitching.
	
	\item \text{Configure camera angle:} Position the camera at 90° (nadir) with de-warping turned off to minimize geometric distortions.
\end{itemize}
\subsubsection{\text{Step 6: Flight Monitoring and Control}}
The operator must continuously monitor battery levels, GPS signal, wind conditions, camera triggering, obstacl-avoidance, and communication stability throughout the flight. Maintain a clear visual line of sight with the UAV at all times, in compliance with aviation safety regulations. Immediately respond to any warnings, errors, deviations to ensure safe operation.
\subsubsection{\text{Step-7 : Conduct Post-Flight Inspection}}
In the post-flight stage firstly, Inspect the drone for any damage or issues immediately after landing and should  record the battery levels of both drone and controller which will help as reference for next flights. Save all mission log files for future planning, troubleshooting, and documenting flight activities. Transfer all captured data to secure storage to prevent loss and ensure it is ready for processing.
\subsubsection{\text{Step 8: Process and Analyze Data}}
Process the raw aerial imagery to generate detailed orthomosaics. Calculate vegetation indices such as NDVI and NDRE to assess crop health. Use the processed data for advanced analysis to identify crop stress and make informed management decisions.

\subsection{\text{Our Procedure}}
Our UAV field data collection procedure follows
\subsubsection{\text{Step 1: Check Airspace Zones}}
\begin{figure}
	\includegraphics[width=1\linewidth]{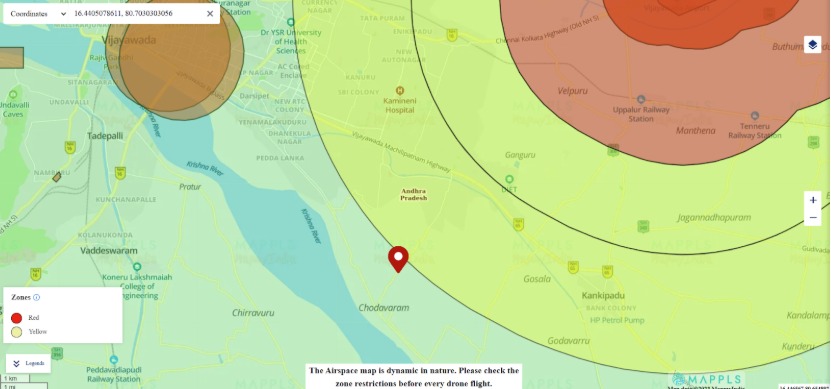}
	\caption{Airspace Map of our field Chodavaram, Vijayawada, Andhra Pradesh in green zone}
	\label{fig:Airspace_map}
\end{figure}

Before each flight, the Digital Sky Airspace Map provided by the Directorate General of Civil Aviation (DGCA), the civil aviation authority of India, was consulted to identify the airspace classification of the survey fields. All sites were verified to comply with DGCA regulations and were located within the green zone	as shown in Figure~\ref{fig:Airspace_map}. 

\subsubsection{\text{Step 2: Check Weather Conditions}}
\begin{figure}
	\centering
	
	\subfloat[Anemometer]{
		\includegraphics[width=0.40\linewidth, height=6cm]{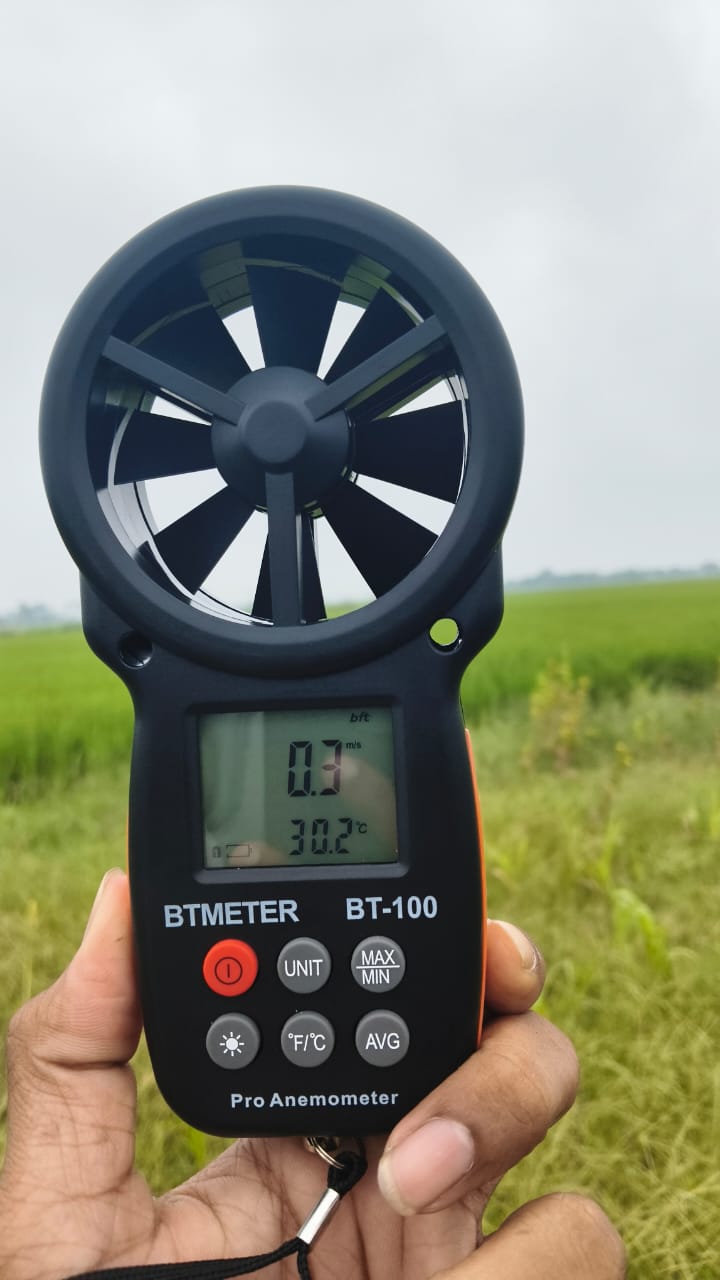}
		\label{fig:anemometer}
	}
	\subfloat[Digital LUX meter]{
		\includegraphics[width=0.40\linewidth, height=6cm]{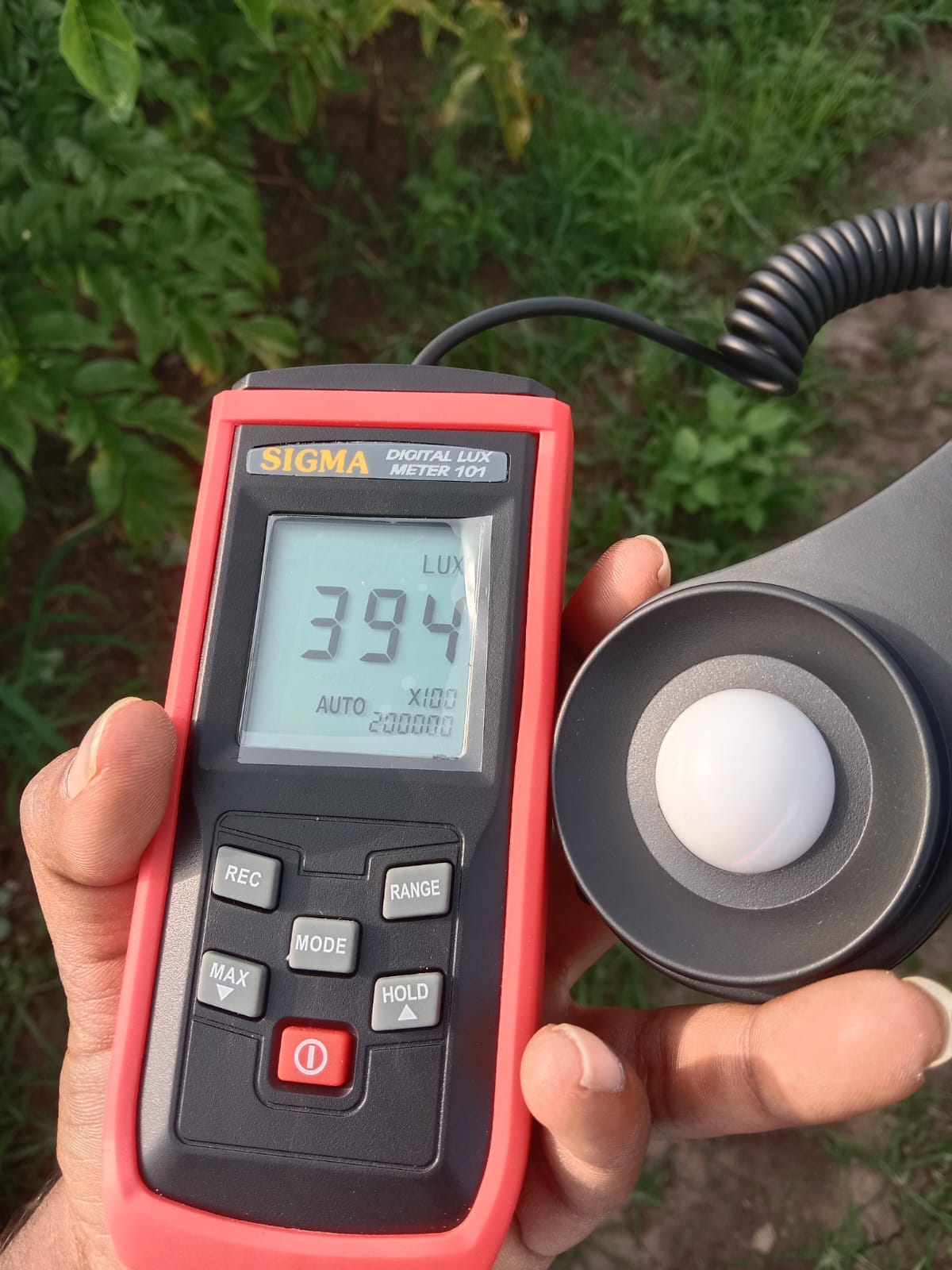}
		\label{fig:luxmeter}
	}
	
	\caption{Measurement Instruments}
	\label{fig:instruments}
\end{figure}

Climatic observations were recorded during each flight mission to ensure accurate environmental context for the captured imagery. Parameters such as temperature, relative humidity, cloud cover, sun elevation and azimuth angles were obtained using the Drone Buddy app, refer Table~\ref{tab:flight_weather}. Wind speed was measured using a handheld anemometer shown in	Figure~\ref{fig:anemometer} oriented in the direction of the wind, while light intensity was recorded by placing a Lux meter shown in Figure~\ref{fig:luxmeter} on a flat surface in the field. These measurements helped account for weather and lighting variations, enhancing the accuracy of orthomosaic generation and vegetation index analyses like NDVI and NDRE.

\begin{table*}
	\centering
	\footnotesize
	\renewcommand{\arraystretch}{1.1}
	\setlength{\tabcolsep}{4pt}
	
	\caption{Flight Missions Weather Data}
	\label{tab:flight_weather}
	
	\begin{tabular}{p{1.4cm} p{1.4cm} p{0.9cm} p{0.9cm} p{0.9cm} 
			p{1.1cm} p{2cm} p{1.1cm} p{1.1cm} p{0.9cm} p{0.9cm}}
		\hline
		\text{Date} & \text{Crop Stage} & \text{Max} & \text{Min} & \text{Wind} &
		\text{Humidity} & \text{Cloud Cover} & \text{Sun Elev.} & \text{Sun Azim.} &
		\multicolumn{2}{c}{\text{Light (LUX)}} \\
		\cline{10-11}
		& & (°C) & (°C) & (m/s) & (\%) & & (°) & (°) & F1 & F2 \\ 
		\hline
		
		15/07/2025 & Nursery & 36 & 27 & 6.9 & 41 & Clear sky & 16.3 & 288.1 & -- & -- \\
		19/07/2025 & Nursery & 35 & 25 & 2.0 & 74 & Sunny & 75.3 & 110.2 & -- & -- \\
		16/08/2025 & Vegetative & 35 & 25 & 4.1 & 74 & Partly cloudy & 49.1 & 257.2 & 149 & 152 \\
		06/09/2025 & Vegetative & 34 & 26 & 1.8 & 66 & Sunny & 19.6 & 90.3 & 345 & 345 \\
		27/09/2025 & Booting & 32 & 24 & 3.3 & 88 & Partly cloudy & 43.3 & 227.1 & 376 & 405 \\
		04/10/2025 & Booting & 36 & 28 & 2.8 & 83 & Clear sky & 29.5 & 250.3 & 430 & 258 \\
		11/10/2025 & Flowering & 29 & 26 & 3.0 & 85 & Partly cloudy & 51.7 & 148.9 & 380 & 440 \\
		21/10/2025 & Flowering & 32 & 26 & 3.5 & 80 & Clear sky & 25.2 & 245.9 & 470 & 400 \\
		08/11/2025 & Mature & 33 & 22 & 2.6 & 75 & Partly cloudy & 8.8 & 253.5 & 360 & 420 \\
		15/11/2025 & Mature & 30 & 19 & 2.6 & 75 & Clear sky & 14.9 & 240.2 & 458 & 380 \\
		\hline
	\end{tabular}
\end{table*}


\subsubsection{\text{Step 3: Optimal Flight Timing}}

Flights were scheduled based on local daylight availability and ambient temperature conditions. Data acquisition was conducted during early morning or late afternoon hours in Vijayawada, Andhra Pradesh, India, as documented by the \emph{Captured Time} column in Tables~\ref{tab:Field1_details} and~\ref{tab:Field2_details}. This scheduling ensured stable flight conditions and consistent image quality across both fields.

\subsubsection{\text{Step 4: Operational Workflow and Flight Execution}}
To ensure safe operations and high-quality data acquisition, we followed a structured workflow covering all flight stages pre-flight, mission planning, in-flight monitoring, and post-flight inspection. 
At each stage, we used the checklists provided below, which were developed based on standard procedures and field-tested parameters.  
During pre-flight and mission planning, we verified all components using this checklist and configured key flight parameters such as GSD- 1~cm, image overlap 80\% forward, 70\% side, and camera angle 90° based on experimental requirements.
During flight, the in-flight checklist helped us monitor drone performance, GPS signal, battery levels, and safety parameters in real time. 
After each mission, we performed a complete post-flight inspection using the post-flight checklist to assess drone condition, save mission logs, and secure captured data. 
This systematic approach ensured both the safety of the drone and the accuracy and reliability of the collected data. We monitored and tabulated the activities taken place in fields such that no event is missed during the data collection provided in Table~\ref{tab:crop_calendar_action}.
We also reviewed several research papers and prepared a comparison table, as summarized in Table~\ref{tab:flight_comparison}. 
			
\begin{table}
	\centering
	\renewcommand{\arraystretch}{1.2}
\setlength{\tabcolsep}{2pt}
				
	\caption{Crop Calendar and Agricultural Activities for the Study Field (5 Acres)}
	\label{tab:crop_calendar_action}
		\begin{tabular}{p{2.5cm} p{1.5cm} p{4cm}}
			\hline
			\text{Stage} & \text{Date} & \text{Activities} \\
			\hline
			Nursery stage    & June 25 & Seed sowed \\
			& July 12 & Seedlings transplanted \\
			& July 16 & Herbicide sprayed \\
			Vegetative stage & Aug 02  & Urea and DAP applied \\
			& Aug 20  & Weeds removed manually \\
			& Sept 05 & NPK fertilizer applied \\
			Booting stage    & Sept 15 & Manual weeding done \\
			& Sept 28 & Fungicide sprayed \\
			Flowering stage  & Oct 10  & Pest monitored \\
			Mature stage     & Oct 25  & Crop matured \\
			Harvesting stage & Nov 29  & Crop harvested \\
			\hline
		\end{tabular}
\end{table}
			

\begin{table}
\caption{Comparison of Flight Operation Parameters in UAV-based Studies}
\label{tab:flight_comparison}
\centering
\scriptsize
\renewcommand{\arraystretch}{1.15}
\setlength{\tabcolsep}{2.5pt}
\begin{tabular}{p{1.6cm} p{0.9cm} p{0.7cm} p{1.2cm} p{1.2cm} p{1.1cm} p{0.9cm}}
	\hline
	Reference &
	\shortstack{Horiz.\\Speed\\(m/s)} &
	\shortstack{Alt.\\(m)} &
	\shortstack{Front\\Overlap (\%)} &
	\shortstack{Side\\Overlap (\%)} &
	\shortstack{GSD\\(cm/px)} &
	\shortstack{Camera\\Mode} \\
	\hline
	~\cite{6_Maulit_2023_Data} & 5   & 57  & 60 & 75 & 3    & Still image \\
	~\cite{7_Fonseka_2024_Brief} & 1.5 & 30  & 80 & 70 & --   & 3 s interval \\
	~\cite{8_Sabine_2024_Brief} & 7   & 80  & 80 & 70 & 4.2--4.6 & -- \\
	~\cite{12_Muhindo_2025_Agrosystems} & 1   & 15  & -- & -- & 1.25 & -- \\
	~\cite{13_Farag_2024_Phenome} & --  & 120 & 75 & 75 & --   & -- \\
	~\cite{15_Wijayanto_2023_AgriEngineering} & 6   & 75  & 80 & 80 & 8.2  & Fast mode \\
	~\cite{17_Govi_2024_Sensing} & 2   & 120 & 80 & 80 & 8    & -- \\
	Our parameters & 3.5 & 22  & 80 & 70 & 1    & Distance interval \\
	\hline
\end{tabular}
\end{table}

\subsubsection{\text{Step 5: Data Processing and Analysis}}

\begin{figure}
	\centering
	
	\subfloat[RGB image]{%
		\includegraphics[width=0.45\linewidth]{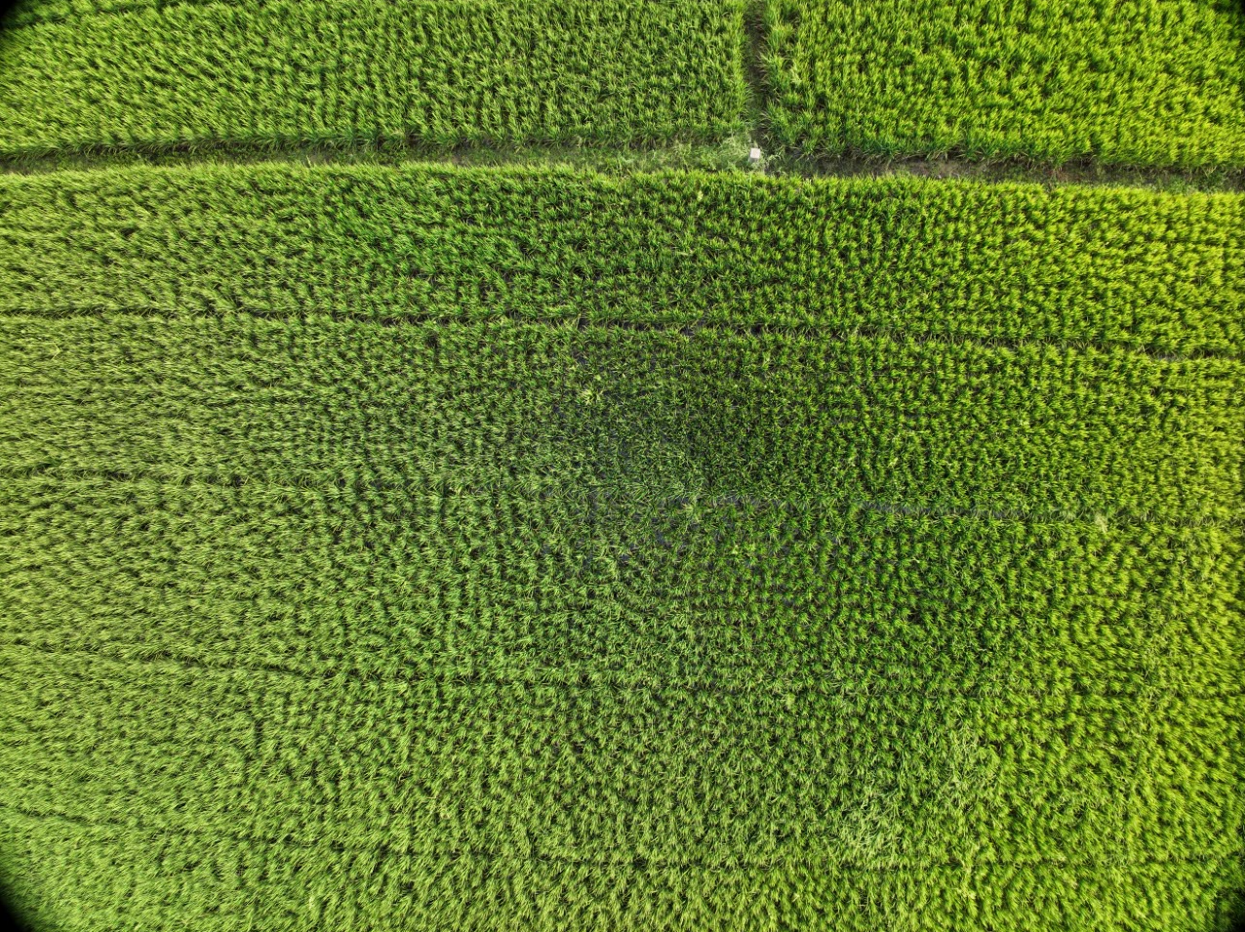}%
		\label{fig:single_rgb}%
	}\hspace{0.04\linewidth}%
	\subfloat[Green band]{%
		\includegraphics[width=0.45\linewidth]{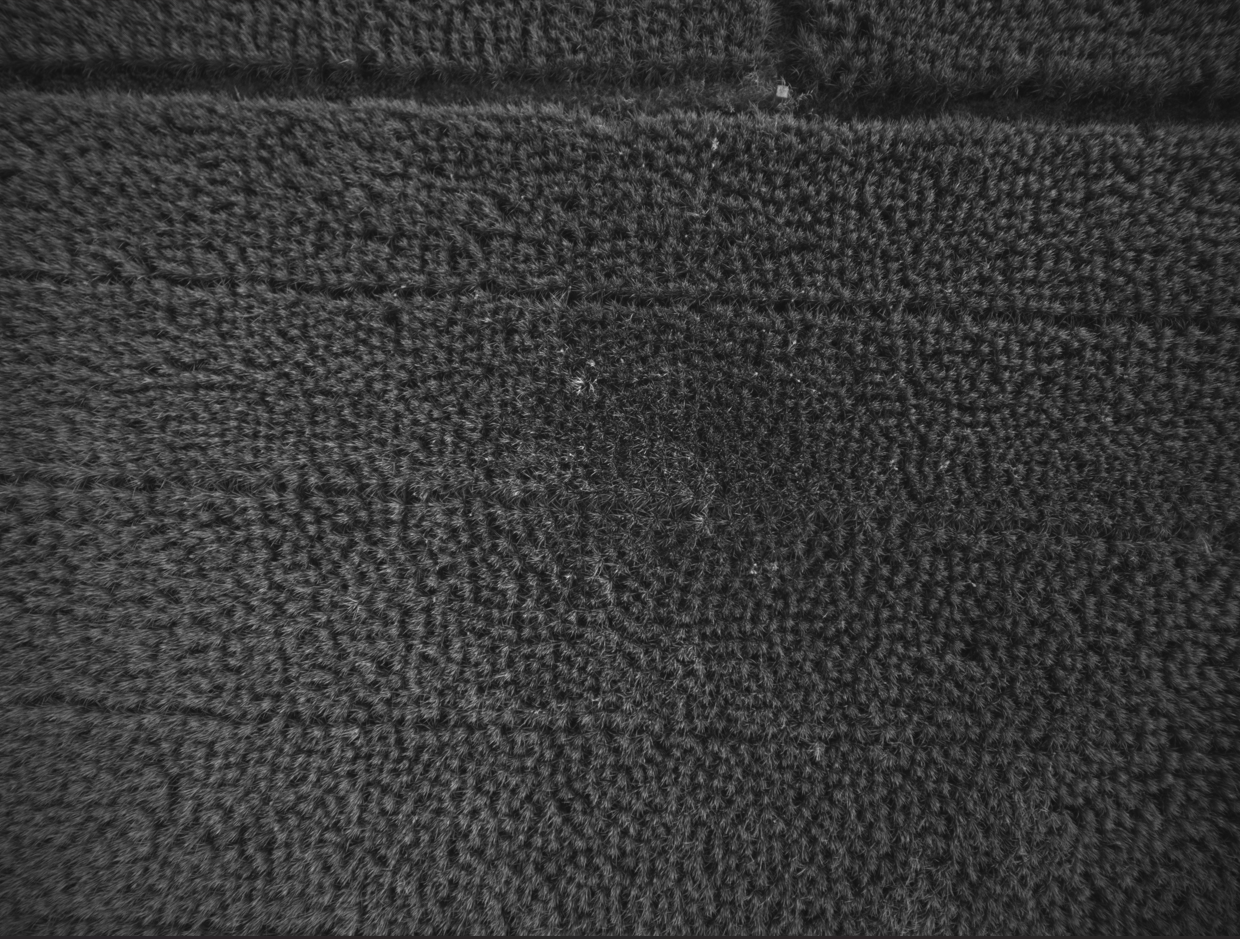}%
		\label{fig:image_green}%
	}
	
	
	\subfloat[NIR band]{%
		\includegraphics[width=0.45\linewidth]{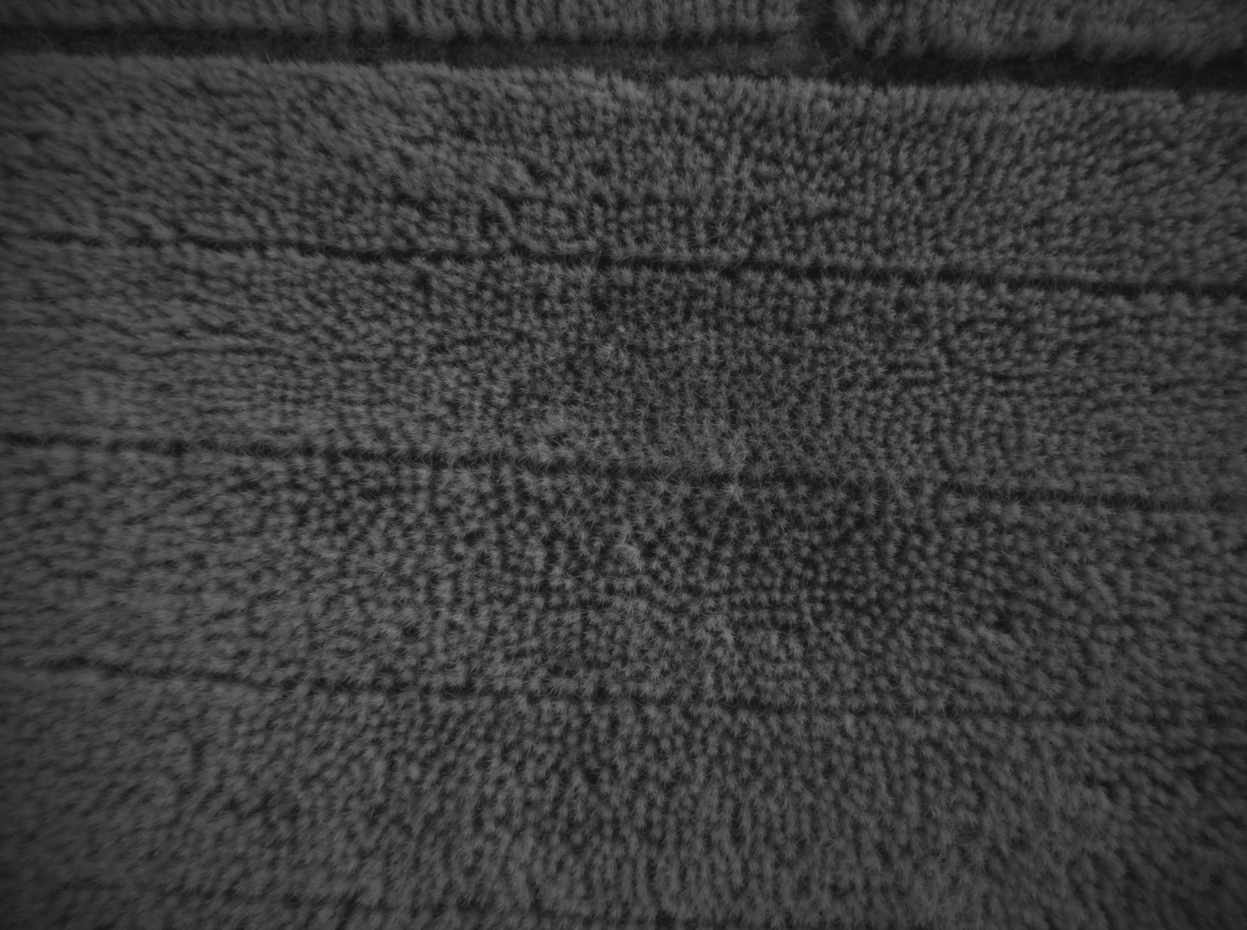}%
		\label{fig:image_nir}%
	}\hspace{0.04\linewidth}%
	\subfloat[Red band]{%
		\includegraphics[width=0.45\linewidth]{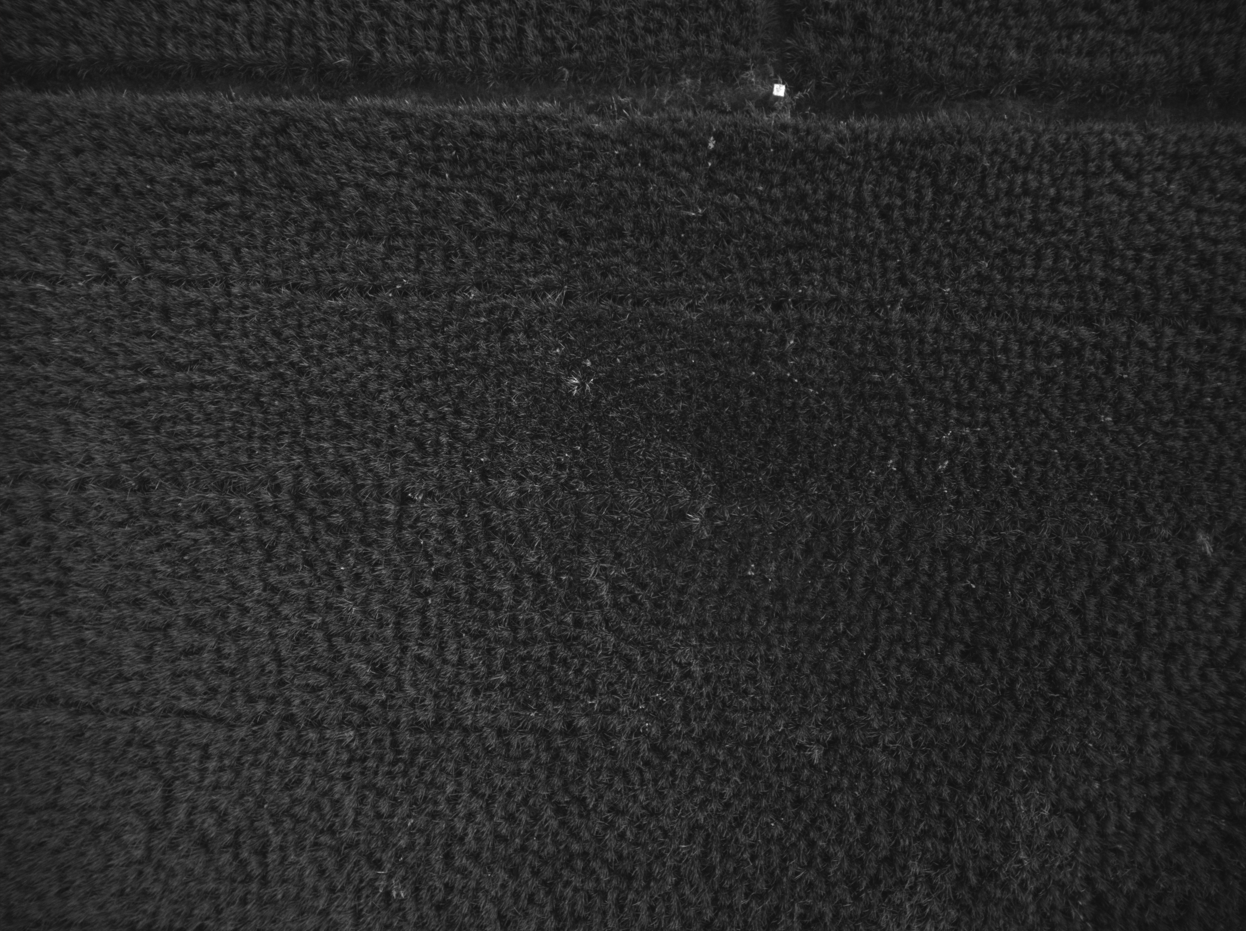}%
		\label{fig:image_red}%
	}
	
	
	\subfloat[RE band]{%
		\includegraphics[width=0.45\linewidth]{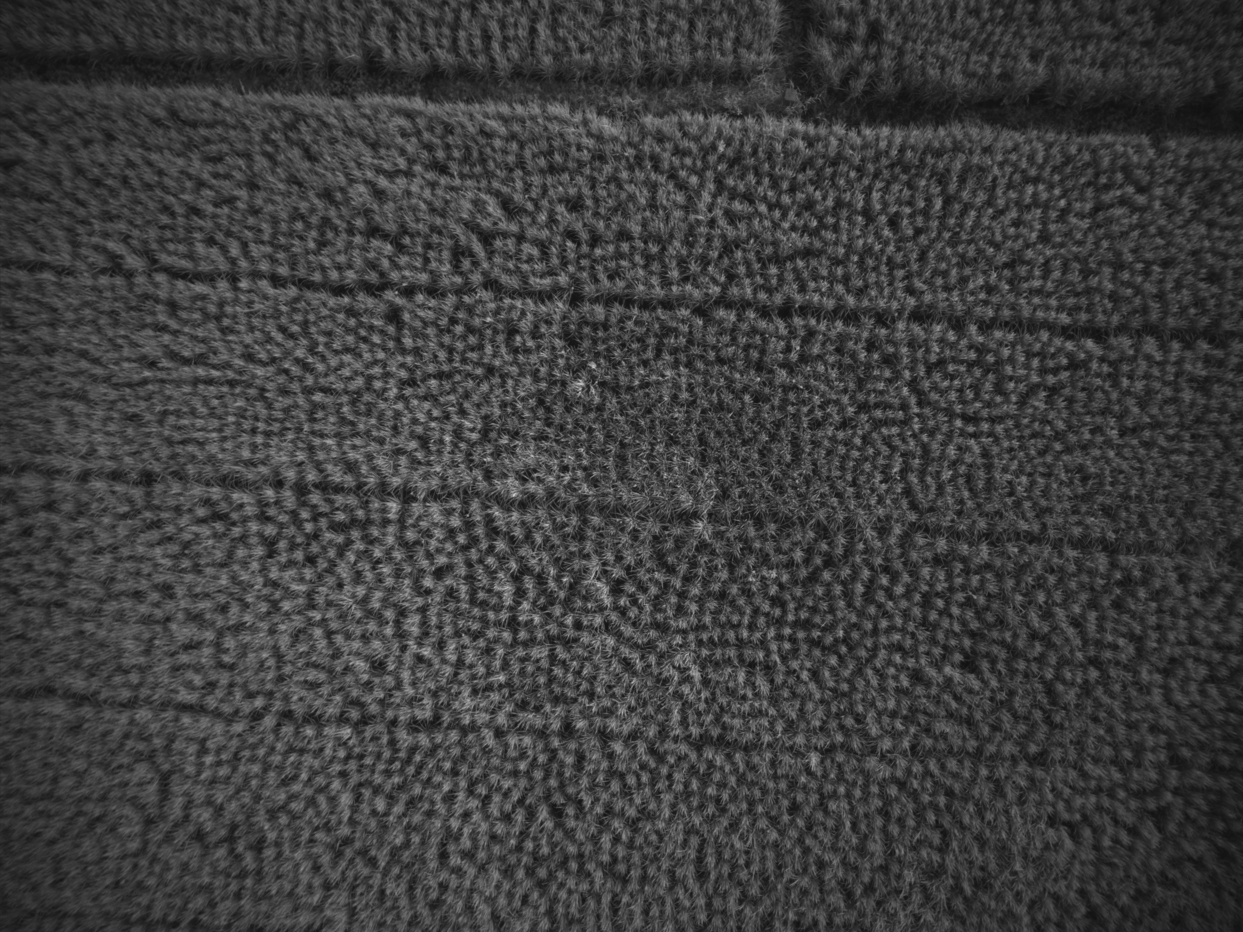}%
		\label{fig:image_re}%
	}
	
	\caption{RGB and multispectral images: (a) RGB, (b) Green, (c) NIR, (d) Red, (e) RE.}
	\label{fig:all_images}
\end{figure}

Raw images shown in Figure~\ref{fig:all_images} were stitched together into orthomosaics. Vegetation indices such as NDVI and NDRE were calculated to assess crop health as shown in Figure~\ref{fig:maps}. The processed data was analyzed to identify crop stress and guide management decisions.

	\renewcommand{\arraystretch}{1.2} 
\begin{table*}
	\caption{Captured Details of Field 1}
	\label{tab:Field1_details}
	\centering
	\footnotesize
	\begin{tabular}{p{1.8cm} p{1.6cm} p{6.4cm} p{1.4cm} p{1.3cm} p{1.3cm} p{1.3cm}}
		\hline
		Date & Captured Time & Folder Name &No.of RGB Images &No.of Multi-spectral Images & Total No. of Images & Storage (GB) \\ \hline
		
		15 Jul 2025 & 17:16 & F1\_01\_2025\_07\_15\_1716\_Nursery\_Stage & 58  & 232  & 290  & 2.61 \\
		19 Jul 2196 & 10:51 & F1\_02\_2025\_07\_19\_1051\_Nursery\_Stage & 549 & 2196 & 2745 & 24.9 \\
		16 Aug 2025 & 16:09 & F1\_03\_2025\_08\_16\_1609\_Vegetative\_Stage & 549 &2196 & 2745& 27.1 \\
		06 Sept 2025 & 07:07 & F1\_04\_2025\_09\_06\_0707\_Vegetative\_Stage & 549 & 2196 & 2745 & 27.5 \\
		27 Sept 2025 & 14:50 & F1\_05\_2025\_09\_27\_1450\_Booting\_Stage   & 549 & 2196 & 2745 & 27.1 \\
		04 Oct 2025 & 15:50 & F1\_06\_2025\_10\_04\_1550\_Booting\_Stage   & 549 & 2196 & 2745 & 27.4 \\
		11 Oct 2025 & 10:48 & F1\_07\_2025\_10\_11\_1048\_Flowering\_Stage & 549 & 2196 & 2745 & 27 \\
		21 Oct 2025 & 16:00 & F1\_08\_2025\_10\_21\_1600\_Flowering\_Stage & 549 & 2196 & 2745 & 27.2 \\
		08 Nov 2025 & 17:00 & F1\_09\_2025\_11\_08\_1700\_Mature\_Stage    & 550 & 2200 & 2750 & 26.6 \\
		15 Nov 2025 & 16:58 & F1\_10\_2025\_11\_15\_1658\_Mature\_Stage    & 549 & 2196& 2745 & 26.5 \\ \hline
		
		Total &  &  & 5000 & 20000 & 25000 & 244 \\ \hline
	\end{tabular}
\end{table*}

\renewcommand{\arraystretch}{1.2} 
\begin{table*}
	\caption{Captured Details of Field 2}
	\label{tab:Field2_details}
	\centering
	\footnotesize
	\begin{tabular}{p{1.8cm} p{1.6cm} p{6.4cm} p{1.4cm} p{1.3cm} p{1.3cm} p{1.3cm}}
		\hline
		Date & Captured Time & Folder Name & No. of RGB Images &No. of Multispectral Images & Total No. of Images & Storage (GB) \\ \hline
		
		15 July 2025 & 17:16 & F2\_01\_2025\_07\_15\_1716\_Nursery\_Stage & 30  & 120  & 150  & 1.32 \\
		19 July 2025 & 09:37 & F2\_02\_2025\_07\_19\_0937\_Nursery\_Stage & 384 & 1536 & 1920 & 16.8 \\
		16 Aug 2025  & 16:39 & F2\_03\_2025\_08\_16\_1639\_Vegetative\_Stage & 384 & 1536 & 1920 & 18.8 \\
		06 Sept 2025 & 06:44 & F2\_04\_2025\_09\_06\_0644\_Vegetative\_Stage & 384 & 1536 & 1920 & 19.1 \\
		27 Sept 2025 & 14:26 & F2\_05\_2025\_09\_27\_1426\_Booting\_Stage   & 384 & 1536 & 1920 & 19 \\
		04 Oct 2025  & 16:23 & F2\_06\_2025\_10\_04\_1623\_Booting\_Stage   & 384 & 1536 & 1920 & 19.3 \\
		11 Oct 2025  & 10:25 & F2\_07\_2025\_10\_11\_1025\_Flowering\_Stage & 384 & 1536 & 1920 & 19 \\
		21 Oct 2025  & 16:46 & F2\_08\_2025\_10\_21\_1646\_Flowering\_Stage & 384 & 1536 & 1920 & 19 \\
		08 Nov 2025  & 16:38 & F2\_09\_2025\_11\_08\_1638\_Mature\_Stage    & 384 & 1536 & 1920 & 18.8 \\
		15 Nov 2025  & 16:37 & F2\_10\_2025\_11\_15\_1637\_Mature\_Stage    & 384 & 1536 & 1920 & 18.8 \\ \hline
		
		Total &  &  & 3486 & 13944 & 17430 & 170 \\ \hline
	\end{tabular}
\end{table*}

\subsection{Standard Checklist} 
We prepared a standardized drone flight checklist and followed it for every flight during data capture, conducted in a standardized manner to ensure safety, consistency, and high-quality outcomes.
\subsubsection{Pre-Flight Checklist}
Prior to each drone mission, a comprehensive pre-flight checklist was followed to ensure operational safety, optimal flight conditions, and high-quality data capture. This included verification of drone and controller battery levels, environmental conditions, flight parameters, and structural integrity, as summarized in Table~\ref{tab:preflight_checklist}.

\begin{table}
	\caption{Drone Flight Pre-Flight Checklist}
	\label{tab:preflight_checklist}
	\centering
	\renewcommand{\arraystretch}{1.3}
	\setlength{\tabcolsep}{3pt}  
	\begin{tabular}{p{3cm} p{2.8cm} p{0.8cm} p{1cm}}
		\hline
		\text{Parameter} & \text{Value/Threshold} & \text{Status} & \text{Remarks} \\
		\hline
		Zone & Red / Green / Yellow & & \\
		Obstacles & Trees, Buildings, Other (Y/N) & & \\
		Drone Battery & $\geq$ 95\% & & \\ 
		Remote Controller Battery & $\geq$ 40\% & & \\ 
		Wind Speed & 5--7 m/s (11--15 mph) & & \\ 
		Satellites Connected & $\geq$ 16 (up to 32) & & \\ 
		Altitude & 30 m & & \\
		Flight Speed & 1.5 m/s (3.35 mph) & & \\
		Shutter Speed & — & & \\
		De-warping & OFF & & \\
		Launch Pad Available & Yes / No & & \\
		Reflectance Panel Used & Yes / No & & \\
		Forward Overlap & 80\% & & \\
		Side Overlap & 70\% & & \\
		Camera Mode & — & & \\
		Camera Orientation & 90° (Nadir) & & \\
		GSD (Ground Sampling Distance) & — & & \\
		Propeller Inspection & No damage, secure & & \\
		Motor Free Spin Test & Spins freely & & \\
		Structural Integrity Check & No cracks/damage & & \\
		Temperature (Min/Max) & — & & \\
		Humidity & — & & \\
		Sun Elevation Angle & — & & \\
		Sun Azimuth Angle & — & & \\
		Sowing Rate (kg/acre) & — & & \\
		\hline
	\end{tabular}
\end{table}

\subsubsection{During-Flight Checklist}
During flight, continuous monitoring of satellite connectivity, battery status, wind conditions, and camera operation was maintained to ensure accurate and uninterrupted data acquisition. Table~\ref{tab:duringflight_checklist} summarizes the key parameters observed throughout each flight.

\begin{table}
	\caption{Drone During-Flight Checklist}
	\label{tab:duringflight_checklist}
	\centering
	\renewcommand{\arraystretch}{1.5}
	\setlength{\tabcolsep}{2.5pt} 
	\begin{tabular}{p{3.2cm} p{2.9cm} p{0.8cm} p{1cm}}
		\hline
		\text{Parameter} & \text{Expected Condition} & \text{Status} & \text{Remarks} \\
		\hline
		Satellite signal & Stable, no drops or losses & & \\
		Battery level monitoring & Sufficient until landing & & \\
		Flight Path Followed & As planned & & \\
		Wind conditions & Acceptable $< 7$ m/s & & \\ 
		Camera trigger & Working, consistent captures & & \\
		Live feed & Stable and visible & & \\
		Communication with drone & No interruptions & & \\
		Avoiding obstacles & Manual override when required & & \\
		\hline
	\end{tabular}
\end{table}

\subsubsection{Post-Flight Checklist}
After landing, a post-flight checklist was performed to confirm the drone’s condition, verify data integrity, and ensure proper storage of equipment and images. Table~\ref{tab:postflight_checklist} presents the parameters and checks completed after each flight mission.

\begin{table}
	\caption{Drone Post-Flight Checklist}
	\label{tab:postflight_checklist}
	\centering
	\renewcommand{\arraystretch}{1.3}
	\setlength{\tabcolsep}{2.5pt} 
	\begin{tabular}{p{3.3cm} p{2.3cm} p{0.8cm} p{1cm}}
		\hline
		\text{Parameter} & \text{Description} & \text{Status} & \text{Remarks} \\
		\hline
		Drone battery level & Record \% & & \\
		Remote battery level & Record \% & & \\
		Propeller re-check & No damage & & \\
		Motor function & No unusual sound & & \\
		Structural check & Frame intact & & \\
		Data downloaded & Yes / No & & \\
		Image quality check & Blurriness / Sharp & & \\
		Control response check & Pitch/Roll/Yaw functional & & \\
		Log saved & Flight data saved & & \\
		Reflectance panel collected & Yes / No & & \\
		\hline
	\end{tabular}
\end{table}

\section{
	Data Description
}

In this study, the UAV data were captured at the Chodavaram, Vijayawada, Krishna District, Andhra Pradesh in India at coordinates [16° 26' 25.8295" N, 80° 42' 10.9091" E]. The survey covered 5 acres with a total of 414 GB of imagery, consisting of 8,486 RGB images and 33,944 multispectral images, making 42,430 images in total. Each dataset included associated metadata such as GPS coordinates, altitude, and time of acquisition.

\subsection{Nomenclature}
\subsubsection{Folder Structure}

The dataset follows a systematic file naming convention for clarity and easy retrieval. We created structured folders to organize the captured images. A primary folder named \text{DATASET} contain subfolders for each surveyed field.

Primary subfolder structure: \text{FieldXXX\_Acre}
\begin{itemize}
	\item XXX is the field number,
	\item Acre is the field area in acres.
\end{itemize}

Example: Field001\_3Acres. 
\par\vspace{0.5em} 
Within each field folder, data is further subdivided into date-time folders structured as:\\
\text{FX\_XX\_YYYY\_MM\_DD\_HHMM\_Stage}
\begin{itemize}
	\item FX corresponds to the field number where F1 for Field001 and F2 for Field002,
	\item XX represents the serial number of that particular field,
	\item YYYY\_MM\_DD represents the captured year, month, and date,
	\item HHMM indicates the hour and minute,
	\item Stage describes the crop growth stage during data collection.
\end{itemize}

Examples:
\begin{itemize}
	\item F1\_01\_2025\_07\_15\_1716\_Nursery\_Stage - data is first survey of Field001 captured on July 15, 2025 at 17:16 during the Nursery Stage.
	\item F2\_03\_2025\_08\_16\_1639\_Vegetative\_Stage - data is third survey of Field002 captured on August 9, 2025 at 17:02 during the Vegetative Stage.
\end{itemize}
\par\vspace{0.5em}

\subsubsection{Image Structure}
\text{RGB Images Format:} \text{YYYYMMDDHHMMSS\_ImageNumber\_D.JPG}
\begin{itemize}
	\item YYYYMMDDHHMMSS: Year, month, date, hour, minute, and second of capture,
	\item ImageNumber: Sequential image number,
	\item D: Indicates downward-facing RGB image.
\end{itemize}

\text{Example:} 20250809173606\_0001\_D.JPG

This file corresponds to the first RGB image captured on August 9, 2025, at 17:36:06.

\vspace{0.5em}

\text{Multispectral Images Format:} \text{YYYYMMDDHHMMSS\_ImageNumber\_Band.TIF}
\begin{itemize}
	\item YYYYMMDDHHMMSS: Year, month, day, hour, minute, and second of capture,
	\item ImageNumber: Sequential image number,
	\item Band: Spectral band designation:
	\begin{itemize}
		\item Green — MS\_G
		\item Red — MS\_R
		\item Red Edge — MS\_RE
		\item Near Infrared — MS\_NIR
	\end{itemize}
\end{itemize}

\text{Example:} 20250809173606\_0001\_MS\_NIR.TIF

This is the first NIR band image captured on August 9, 2025, at 17:36:06.

\subsection{
	Data Organization
}

\begin{figure}
	\centering
	\includegraphics[width=0.98\linewidth, height=13.5cm]{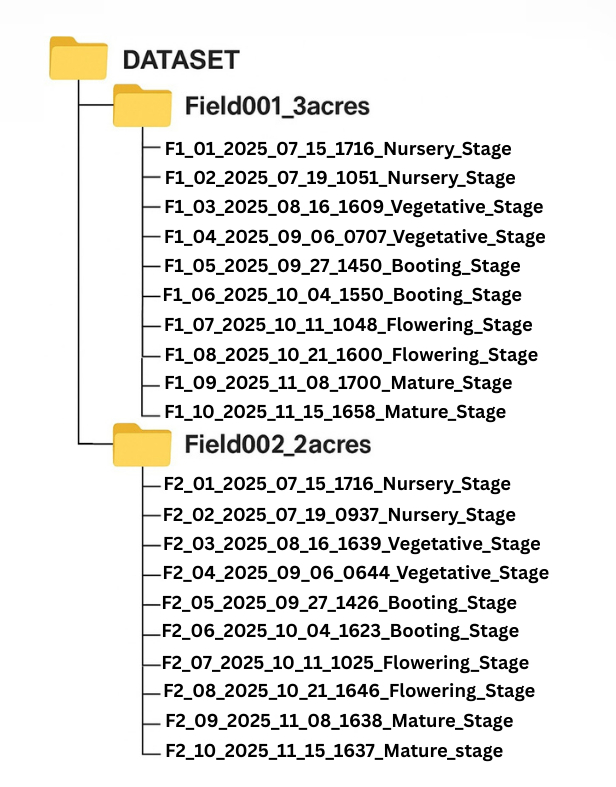}
	\caption{Folder Structure}
	\label{fig:folder_struc}
\end{figure}

We created structured folders to manage the captured images. A primary folder named \text{Dataset} contains subfolders for each surveyed field. Each field folder is named using the format \text{FieldXXX\_Size}. 

Each field folder is further organized into four subfolders based on the capture dates and time: 
\\\text{FX\_XX\_YYYY\_MM\_DD\_HHMM\_Stage} as shown in Figure~\ref{fig:folder_struc}. 

Within each dated subfolder, there are three components: system files, RGB images (.JPG format), and multispectral images (.TIF format) comprising Green, Red, NIR, and RE bands.		
This hierarchical structure ensures clear separation of datasets by field and date, while also simplifying retrieval and analysis. The details of captured images and storage for each field are summarized in Table~\ref{tab:Field1_details} and Table~\ref{tab:Field2_details}.
The current release corresponds to dataset version v1.0, with future updates planned based on additional field campaigns.

	\begin{figure*}
	\centering
	
	\subfloat[]{
		\includegraphics[width=0.40\linewidth, height=6cm]{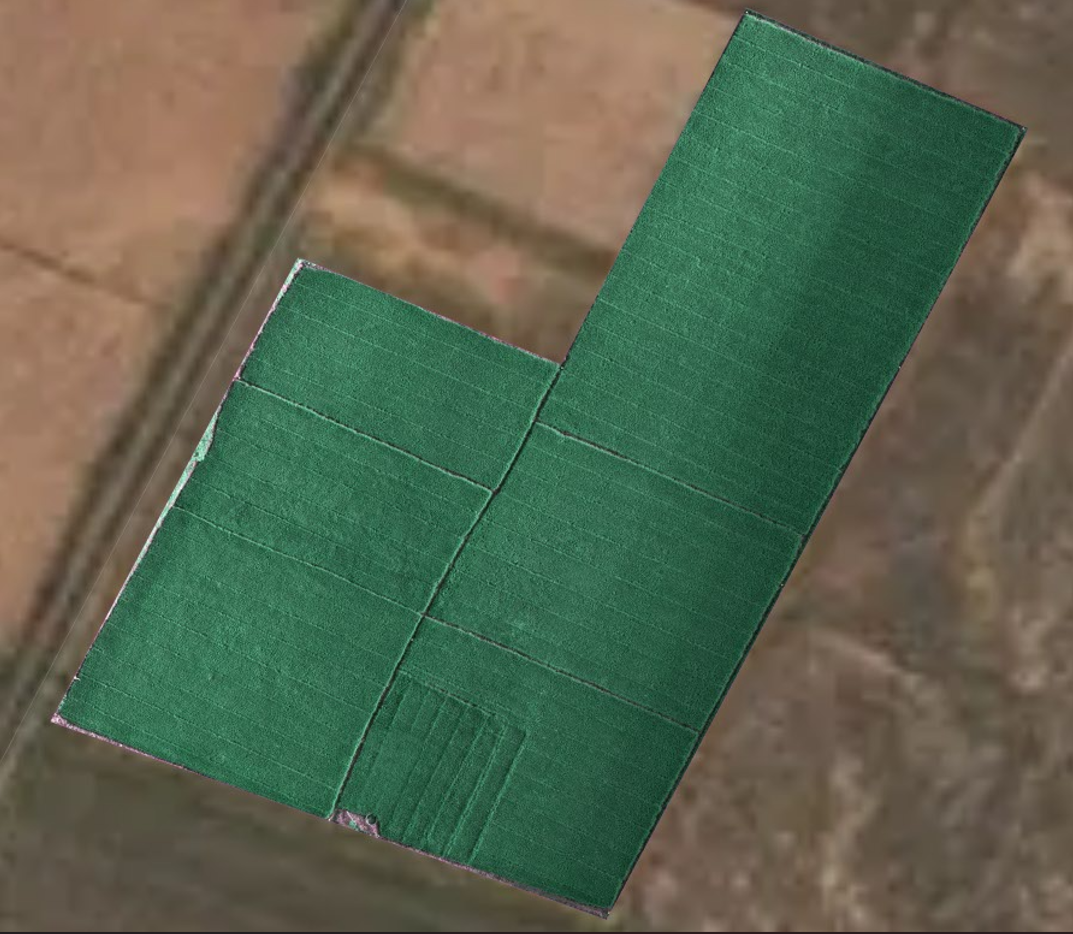}
	}\hspace{0.05\linewidth}
	\subfloat[]{
		\includegraphics[width=0.40\linewidth, height=6cm]{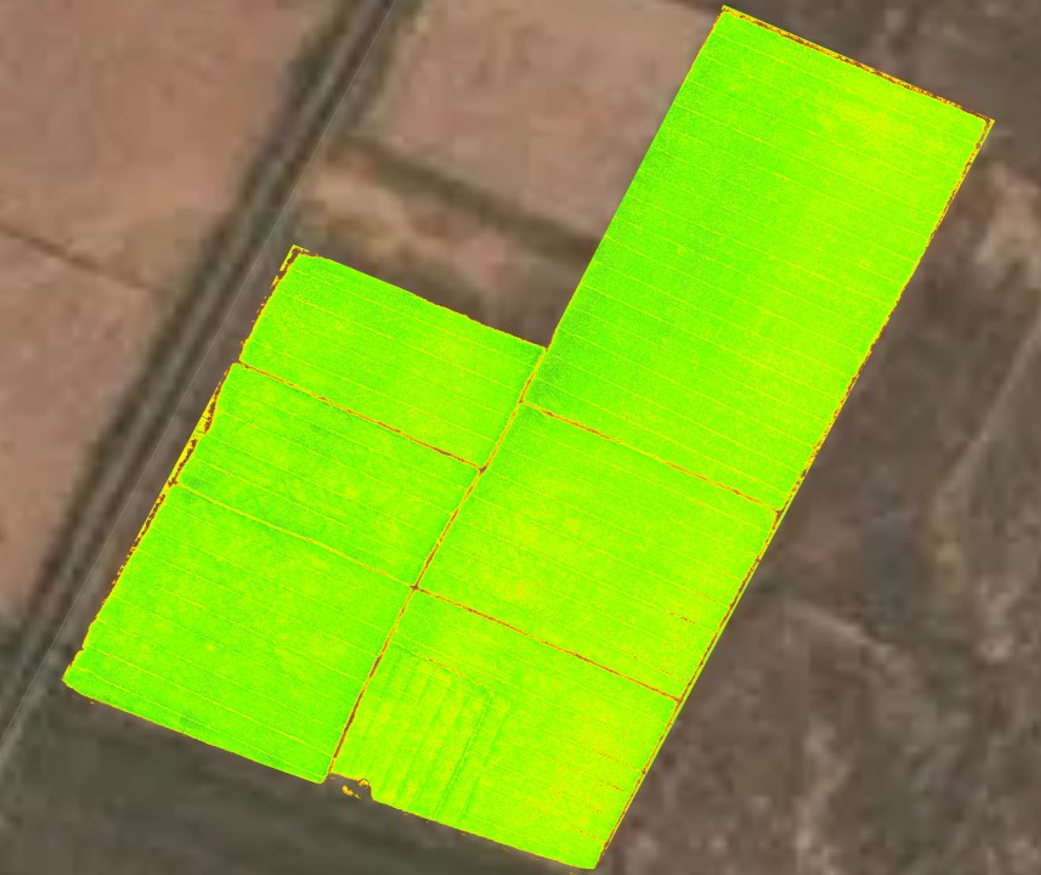}
	}
	
	\subfloat[]{
		\includegraphics[width=0.40\linewidth, height=6cm]{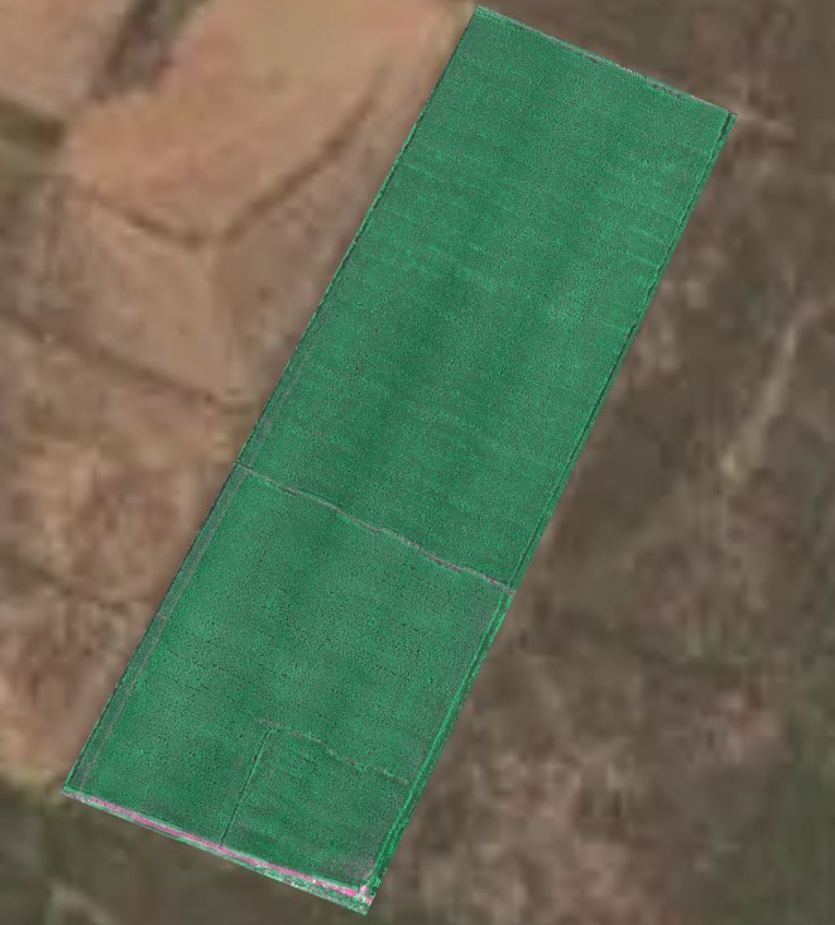}
	}\hspace{0.05\linewidth}
	\subfloat[]{
		\includegraphics[width=0.40\linewidth, height=6cm]{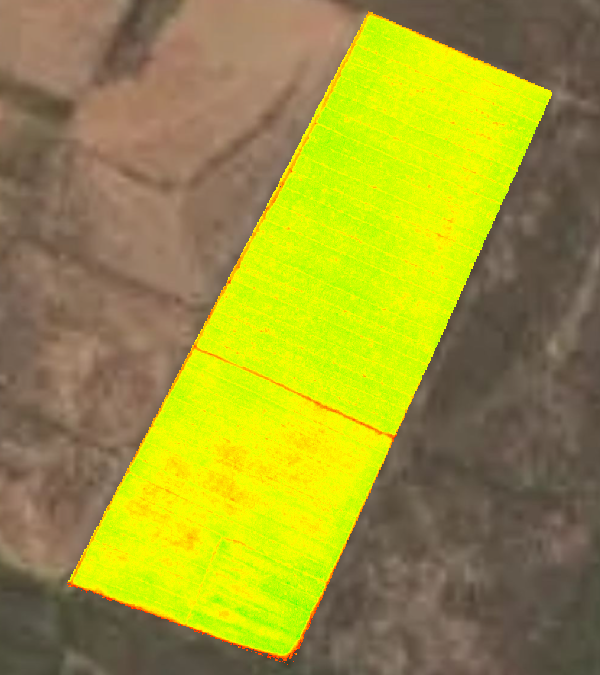}
	}
	
	\caption{
		Visualization of orthomosaic and NDVI outputs:  
		(a) Field001 – 3 Acres Orthomosaic,  
		(b) Field001 – 3 Acres NDVI,  
		(c) Field002 – 2 Acres Orthomosaic,  
		(d) Field002 – 2 Acres NDRE.
	}
	\label{fig:maps}
\end{figure*}

\section{Data Quality and Validation}
Data quality and validation were ensured through systematic checks during image acquisition and processing. Radiometric consistency was maintained by capturing imagery under comparable illumination conditions and applying sensor-based calibration within the photogrammetry workflow. Adequate forward and side image overlap was verified to ensure reliable feature matching and stable orthomosaic reconstruction. Embedded GPS metadata were examined to maintain positional consistency across images, and the generated orthomosaics were evaluated for spatial continuity and alignment accuracy, with visual inspection used to identify and minimize residual stitching or alignment errors. Temporal consistency was preserved by organizing the dataset according to predefined paddy growth stages and maintaining uniform acquisition protocols across all flight campaigns. 
\subsubsection{Pix4D Software}

Pix4D Fields version 2.19, operated under a valid license, was used to process the aerial imagery collected during the experiment. The software was employed to generate high-resolution orthomosaic maps and vegetation index layers, including NDVI and NDRE, as illustrated in Figure~\ref{fig:maps}. The processing workflow involved georeferencing individual images, stitching them into seamless orthomosaics, and computing vegetation indices to assess crop health and spatial variations in plant vigor and stress.		
From the dataset, one folder from Field 1 (3 acres) named F1\_03\_2025\_08\_16\_1609 containing 2,196 multispectral images, and another folder from Field 2 (2 acres) named F2\_03\_2025\_08\_16\_1639 containing 1,920 multispectral images were processed using Pix4D Fields. Two separate orthomosaics were generated, and the NDVI layers were applied to classify crop health. The resulting outputs, including orthomosaic, NDVI, and NDRE maps for both Field 1 and Field 2, are shown in the Figure~\ref{fig:maps}.

%
%
%
%
%
%
%

\subsection{Use Cases of the Dataset}

\begin{itemize}
	\item Our work facilitates the SOP and checklists for drone-based agricultural monitoring.
	
	\item Supports UAV mission planning and optimization, including flight parameters, image overlap, acquisition timing, and high-resolution imagery at 1~cm/pixel GSD for precise vegetation index assessment.
	
	\item Offers insights for agricultural policy and extension services, supporting data-driven decision-making and sustainable crop management.
	
	\item Our dataset can  transfer learning, allowing models trained on other crops or regions to be adapted to tropical Indian paddy fields.
	
	\item Provides a large-scale, publicly available UAV dataset covering RGB and multispectral imagery across all paddy growth stages, enabling comprehensive research and analysis.
	
	\item Precision crop monitoring and management: High-resolution RGB and multispectral imagery, combined with NDVI and NDRE layers, can support targeted spraying, early pest and stress detection, weed and disease analysis, and accurate crop classification.
	
	\item Temporal analysis and modeling: Complete coverage from nursery to harvest supports growth-stage monitoring, biomass estimation, yield prediction, and temporal change assessment, facilitating advanced crop modeling and predictive analytics.
	
\end{itemize}

\section{Conclusion}

Precision agriculture faces challenges in effectively monitoring crops throughout their full growth cycle, as variability in growth stages and environmental conditions can affect yield. To address this, we developed a large-scale UAV-based RGB and multispectral dataset over 5 acres with high-resolution imagery at 1~cm/pixel GSD, covering the complete paddy growth cycle from nursery to harvest stages. This dataset can support accurate monitoring, stress detection, and growth-stage analysis. By following a standardized operating procedure for UAV flights and image processing, we ensured consistent, high-quality data acquisition and reliable metadata collection. This work fills a critical gap in publicly available datasets for paddy crops and supports precision agriculture applications such as yield optimization, crop classification, vegetation index analysis, and machine learning–based predictive modeling. The dataset serves as a valuable resource for researchers and practitioners aiming to enhance crop monitoring practices and improve decision-making in paddy cultivation.



		
\end{document}